%% file: main.tex
\documentclass[conference]{IEEEtran}
\IEEEoverridecommandlockouts
\usepackage{cite}
\usepackage{svg}
\usepackage{amsmath,amssymb,amsfonts}
\usepackage{algorithmic}
\usepackage{graphicx}
\usepackage{textcomp}
\usepackage{xcolor}
\usepackage{url}

\def\BibTeX{{\rm B\kern-.05em{\sc i\kern-.025em b}\kern-.08em
    T\kern-.1667em\lower.7ex\hbox{E}\kern-.125emX}}

\begin{document}

\title{MPRU: Modular Projection–Redistribution Unlearning as Output Filter for Classification Pipelines\\
}

\author{\IEEEauthorblockN{1\textsuperscript{st} Minyi PENG}
\IEEEauthorblockA{\textit{Nanyang Technological University} \\
Singapore \\
minyi002@e.ntu.edu.sg}
\and
\IEEEauthorblockN{2\textsuperscript{nd} Darian Gunamardi}
\IEEEauthorblockA{\textit{Nanyang Technological University} \\
Singapore \\
darian.gunamardi@ntu.edu.sg} 

\and
\IEEEauthorblockN{3\textsuperscript{rd} Ivan Tjuawinata}
\IEEEauthorblockA{\textit{Nanyang Technological University} \\
Singapore \\
ivan.tjuawinata@ntu.edu.sg}


\and
\IEEEauthorblockN{4\textsuperscript{th} Kwok-Yan Lam}
\IEEEauthorblockA{\textit{Nanyang Technological University} \\
Singapore \\
kwokyan.lam@ntu.edu.sg}
}

\maketitle

\begin{abstract}


As a new and promising approach, existing machine unlearning (MU) works typically emphasize theoretical formulations or optimization objectives to achieve knowledge removal. However, when deployed in real-world scenarios, such solutions typically face scalability issues and have to address practical requirements such as full access to original datasets and model. 



In contrast to the existing approaches, we regard classification training as a sequential process where classes are learned sequentially, which we call \emph{inductive approach}. Unlearning can then be done by reversing the last training sequence. This is implemented by appending a projection-redistribution layer in the end of the model. Such an approach does not require full access to the original dataset or the model, addressing the challenges of existing methods.
%
This enables modular and model-agnostic deployment as an output filter into existing classification pipelines with minimal alterations. 



We conducted multiple experiments across multiple datasets including image (CIFAR-10/100 using CNN-based model) and tabular datasets (Covertype using tree-based model). Experiment results show consistently similar output to a fully retrained model with a high computational cost reduction. This demonstrates the applicability, scalability, and system compatibility of our solution while maintaining the performance of the output in a more practical setting.

\end{abstract}


\section{Introduction}




Machine unlearning enables models to forget previously learned information, and this process is typically approached in two ways: retraining and parameter-tuning. Retraining, following exact unlearning, treats naive retraining as the gold standard, requiring the unlearned model to match the retrained model’s distribution and erase all information about the forget set. 
In general, retraining approaches are thoroughly complete with a stronger theoretical guarantee. However, it comes with several challenges including considerable computational cost \cite{amnesia} and the need of access to the original dataset \cite{SISA}, making such approaches challenging to deploy in real life systems.

Parameter-tuning, an approximate unlearning approach, iteratively adjusts model weights, often via gradient updates on small datasets.
It is suitable in small scales, but shows diminishing performance and adaptation challenges in large or complex models\cite{survey}.
Furthermore, its verification is fragile due to over-reliance on accuracy or attack-based metrics \cite{zhang2024verificationmachineunlearningfragile}, and its best outcome still depends on retraining as a reference.

In terms of data, the coverage of removed concept has spread from individual points to groups, with this work focusing on \emph{class-level removal}. When the forget set corresponds to a coherent semantic category, class-level unlearning cleans the decision boundary more effectively than point-wise erasure \cite{boundeary_unlearning}. Existing approaches fall into instance-level and feature-level categories \cite{instance_unlearning, feature_unlearning}. Instance-level methods modify datasets through pruning, obfuscation, or replacement \cite{SISA, data_obfuscation, data_replacement}, which is incompatible with systems of transient data or restricted access. Feature-level methods restructure entire workflows or alter parameters (e.g., shifting, pruning) \cite{SISA, certified_data_removal, randomised_trees_for_low-latency_MU, federeated_class_discriminative_pruning}, requiring infrastructure changes and facing similar challenges as gradient-based parameter-tuning.

In contrast to the existing approaches, we consider a different direction. First, we regard the training of a classifier as a step-by-step approach where classes are learned sequentially, which we call \emph{inductive approach}. To unlearn a class, we can see the learning of such a class as the last training sequence of such training process. Unlearning can then be considered as reversing this last training sequence. 

In this work, we introduce MPRU, which is a process to reverse the last training sequence. Our approach processes the output of the original model via two steps: projection and redistribution in the attempt to reverse the learning of the label to be forgotten. This process is then appended to the end of the original model as a modular output filter to produce the desired output. Such an approach does not require knowledge about the model structure or modification of the existing parameter and enable integration of the new model into existing classification pipelines with minimal changes. Hence MPRU provides a machine unlearning solution that is model-agnostic, has low complexity, requires little alterations to the original model workflow, and only requires access to the output of the original model given testing data as input without further access to either the training data or the original model.

%
%
We demonstrate its applicability across domains by running experiments using image (CIFAR-10/100) and tabular (Covertype) datasets, with CNN-based ResNet and tree-based XGBoost models respectively. Consistent performance on large-scale Covertype data further highlights scalability and deployment readiness.

In summary, our contributions are the following:

\begin{itemize}
    \item We consider inductive approach in class unlearning that enables the simulation of classification models as reversible systems (See Section \ref{sec:inductivity}).
    \item Based on this approach, we design MPRU, a projection-redistribution unlearning scheme, whose operation only requires model output. This allows modular deployment as a model-agnostic output filter (See Section \ref{sec:proposedsol}).
    \item We present a detailed analysis to show its scalability and low complexity (See Section \ref{sec:complexity}), backed with time comparison results. 
    \item We demonstrate its applicability through experiments using image and tabular datasets, CNN-based and tree-based models, and bound our performance with accuracy, KL-divergence, and MSE (See section \ref{sec:experiment}).
\end{itemize}

\section{Preliminary}
\subsection{Notations}

Given a function $f:A\rightarrow B$ and $S\subseteq A,$ we define $f(S)=\{f(a):a\in S\}\subseteq B.$

For any statement $\mathfrak{s},$ we define the indicator function $\mathbb{I}(\cdot)$ such that $\mathbb{I}(\mathfrak{s})$ returns $1$ if the statement $\mathfrak{s}$ is correct and $0$ otherwise.

Given a real-valued vector $\mathbf{v}$ of length $n,$ for $p=1,2,$ define its $L_p$-norm by 
\[\|\mathbf{v}\|_p\triangleq \left(\sum_{i=1}^n |v_i|^p\right)^{1/p}.\]

Lastly, for a non-zero non-negative vector $\mathbf{v}=(v_1,\cdots, v_a)\in \mathbb{R}^a_{\geq 0}$ of length $a,$ we say $\tilde{\mathbf{v}}=(\tilde{v}_1,\cdots, \tilde{v}_a)\in \mathbb{R}^a_{\geq 0}$ is its \emph{$L_1$-normalization} or \emph{sum-to-one normalization} if
\[\tilde{\mathbf{v}}\triangleq \frac{\mathbf{v}}{\|\mathbf{v}\|_1},\]
which ensures that $\sum_{i=1}^a \tilde{v}_i=1.$

\subsection{General Class Unlearning Scheme and Evaluation Metrics}\label{sec:general_scheme_metrics}

Let $\mathcal{X}$ be a data space and $\mathcal{L}$ be a label space of size $n$. For simplicity, without loss of generality, we assume that $\mathcal{L}=\{1,\cdots, n\}.$ Consider a labeled dataset $\mathcal{Z}=\{\mathbf{z}_1,\cdots, \mathbf{z}_N\}\subseteq \mathcal{X}\times \mathcal{L}$ where we denote $\mathbf{z}_i=(x_i,y_i)$ for $x_i\in \mathcal{X}$ and $y_i\in \mathcal{Y}.$ Without loss of generality, we may assume that $x_i\neq x_j$ for any $i\neq j.$ Denote by $\mathcal{D}=\{x_1,\cdots, x_N\}\in \mathcal{X}$ where for $x\in \mathcal{D},$ denote by $y_{x},$ its corresponding label. For $i=1,\cdots, N,$ denote by $\mathcal{D}_i=\{x\in \mathcal{D}:y_x=i\}.$ This partitions $\mathcal{D}$ to $n$ subsets based on its label.

For a class unlearning scenario, we assume the existence of $\mathcal{L}_u\subseteq \mathcal{L}$ the set of labels we would like to unlearn of size $n_u$. 
Define $\mathcal{D}_u\triangleq\{x\in \mathcal{D}:y_x\in \mathcal{L}_u\},$ the set of data points with label to be unlearned of size $N_u$ which we call \emph{forget set}. Similarly, we define $\mathcal{L}_r\triangleq \mathcal{L}\setminus \mathcal{L}_u,$ the set of labels to be retained of size $n_r$ and $\mathcal{D}_r \triangleq \mathcal{D}\setminus \mathcal{D}_u,$ the set of data points with label to be retained of size $N_r$ which we call \emph{retain set}. We denote the corresponding labeled retained dataset to be $\mathcal{Z}_r.$

We define $M^P:\mathcal{X}\rightarrow [0,1]^{|\mathcal{L}|},$ a classifier trained using the whole $\mathcal{D}$ with labels from $\mathcal{L}.$ Here for any $x\in \mathcal{X},$ we assume that $M^{P}(x)=(r_1,\cdots, r_n)$ where  $\sum_{i=1}^n r_i=1.$ Here for any such a machine $M'$ with label set $\mathcal{L}'$ and an input $x\in \mathcal{X},$ we call the vector $\mathbf{c}_x=M(x)$ the \emph{confidence vector of $M'$ given input $x$ over label set $\mathcal{L}'$}. Given $M^P$ and $\mathcal{L}_u$, the problem of \emph{unlearning $\mathcal{L}_u$ from $M^P$} aims to output a new classifier $M^U:\mathcal{X}\rightarrow [0,1]^{|\mathcal{L}_r|}.$

We note that a straightforward way for unlearning $\mathcal{L}_u$ can be done by simply retraining an entirely new classifier $M^R:\mathcal{X}\rightarrow[0,1]^{|\mathcal{L}_r|}$ trained using $\mathcal{Z}_r.$ Here $M^R$ is the retrained model, which can be seen as the target model we would like to achieve without entire retraining from scratch.

In order to evaluate the performance of $M^U,$ we compare it with $M^R.$ Intuitively, we require $M^U$ and $M^R$ to have similar behavior. This can be measured through the distribution of the respective outputs. This can be measured through the Kullback-Leibler Divergence (KL divergence) (reference for example \cite{Csi75}) between the outputs of the two machines given the retain set and unlearn set. More specifically, for $\delta_r,\delta_u>0,$ we say that $M^U$ is \emph{statistically within} $(\delta_r,\delta_u)$ to $M^R$ if 


\[
\begin{array}{c}
   KL(M^U(D_r)\|M^R(D_r))\leq \delta_r\\  \mathrm{~and~} \\
      KL(M^U(D_u)\|M^R(D_u))\leq \delta_u
\end{array}
\]

Alternatively, we may simplify the evaluation further. First, for any set of labels $\mathcal{L}'\subseteq \mathcal{L}$ and a machine $M:\mathcal{X}\rightarrow [0,1]^{|\mathcal{L}'|},$ for any $x\in \mathcal{X},$ we define $\hat{y}_x,$ the predicted label from $M,$ That is,
\[\hat{y}_{x}= \hat{M}(x)\triangleq \arg\max M(x).\]
We further assume that the true label of $x$ is $y_x\in \mathcal{L}.$

Given such a predicted label, we can measure the accuracy of any machine $M'$ with respect to $\mathcal{D}_r$ by
\[
\mathtt{Accuracy}(M',\mathcal{D}_r)=\frac{1}{N_r}\sum_{x\in \mathcal{D}_r}\mathbb{I}(\hat{M'}(x)=y_x).
\]

The performance $M^U$ can then be measured by the similarity of the accuracies of $M^U$ and $M^R$ with respect to $\mathcal{D}_r.$ More specifically, for $\epsilon_r>0,$ we say that $M^U$ achieves $\epsilon_r$ \emph{accuracy difference} to $M^R$ if 
\[\left|\mathtt{Accuracy}(M^U,\mathcal{D}_r)-\mathtt{Accuracy}(M^R,\mathcal{D}_r)\right|<\epsilon_r.\]

Next, we note that we would also like the performance of $M^U$ to not deteriorate too much from the performance of $M^P,$ in particular with respect to $\mathcal{D}_r.$ Hence, for $\epsilon_p>0,$ we say that $M^U$ achieves $\epsilon_p$ \emph{accuracy difference} to $M^P$ if
\[\left|\mathtt{Accuracy}(M^U,\mathcal{D}_r)-\mathtt{Accuracy}(M^P,\mathcal{D}_r)\right|<\epsilon_p.\]

Next, performance similarity when considering data with labels in $\mathcal{L}_u$ should not be measured based on the highest probability in the output. This is because in either case, the predicted label will be wrong. Because of this, instead, we can consider the mean squared error of the outputs. More specifically, we define
\[\mathtt{MSE}(M^U,M^R,\mathcal{D}_u)\triangleq\frac{1}{N_u}\sum_{x\in \mathcal{D}_u} \|M^U(x)-M^R(x)\|_2^2.\]

For $\epsilon_u>0,$ we say that $M^U$ achieves $\epsilon_u$ \emph{unlabelled-similarity} to $M^R$ if $\mathtt{MSE}(M^U,M^R,\mathcal{D}_u)<\epsilon_u.$

Lastly, we are also interested in the efficiency of the unlearning method. Here, we measure such efficiency by complexity and the time it takes for calculating $M^U(x)$ given $M^P$ and $x\in \mathcal{X}.$ 

\section{Problem Statement}




In this work, we consider the following class unlearning problem. Let $M^P:\mathcal{X}\rightarrow[0,1]^{n}$ be a model that has been trained using a dataset $\mathcal{D}^T$ with labels in $\mathcal{L}=\{1,\cdots, n\}.$  Let $\mathcal{D}$ be the test data that is used for the unlearning process. Assume that we would like to unlearn one of the labels $j\in \{1,\cdots, n\}.$ That is, $\mathcal{L}_u=\{j\}$ and $\mathcal{L}_r=\mathcal{L}\setminus\{j\}.$ We aim to define a protocol $\Pi$ such that given $M^P$ and $\mathcal{L}_u,$ it produces the unlearned model $M^U:\mathcal{X}\rightarrow [0,1]^{n-1}.$ Furthermore, we evaluate the performance of $\Pi$ by finding the values of $\delta_r,\delta_u,\epsilon_r,\epsilon_p,\epsilon_u$ such that $M^U$ is $(\delta_r,\delta_u)$ statistically similar to $M^R$ while having $\epsilon_r$ accuracy similarity, $\epsilon_p$-accuracy retention, and $\epsilon_u$ unlabeled similarity to $M^R.$ Here we define $\mathcal{D}_u$ and $\mathcal{D}_r$ as subsets of $\mathcal{D}$ corresponding to the forgotten label and the retained labels respectively.

\subsection{Assumptions}\label{assumptions}

In this work, we assume some performance guarantee on $M^P.$ As defined before, for any machine $M$ that outputs confidence vector, we can define a corresponding machine $\hat{M}$ that outputs prediction by taking the label corresponding to the highest confidence level. We require that the accuracy of $\hat{M}^P$ in $\mathcal{D}$ to be at least $0.8.$ More specifically, we have that for any $i=1,\cdots, n$ and for any $x\in \mathcal{D}_i, \mathrm{Pr}(\hat{M}^P(x)=i)\geq 0.8$ where the randomness is taken over the choice of $x.$ Furthermore, we require that for any $i=1,\cdots, n$ and $x\in \mathcal{D}_i,$ if $\hat{M}^P(x)=i,$ then $\max(M^P(x))\geq 0.7.$ These constrain the variance of predicted confidence in $M^P$. By the assumptions we have made, we can assume that the confidence vectors output by $M^P$ for such data points have a smaller variance around their averages. This provides a sharper estimate of the projection space.

\subsection{Discussion on Inductivity in Learning}\label{sec:inductivity}

\begin{figure}[h]
    \centering
    \includegraphics[width=0.48\textwidth]{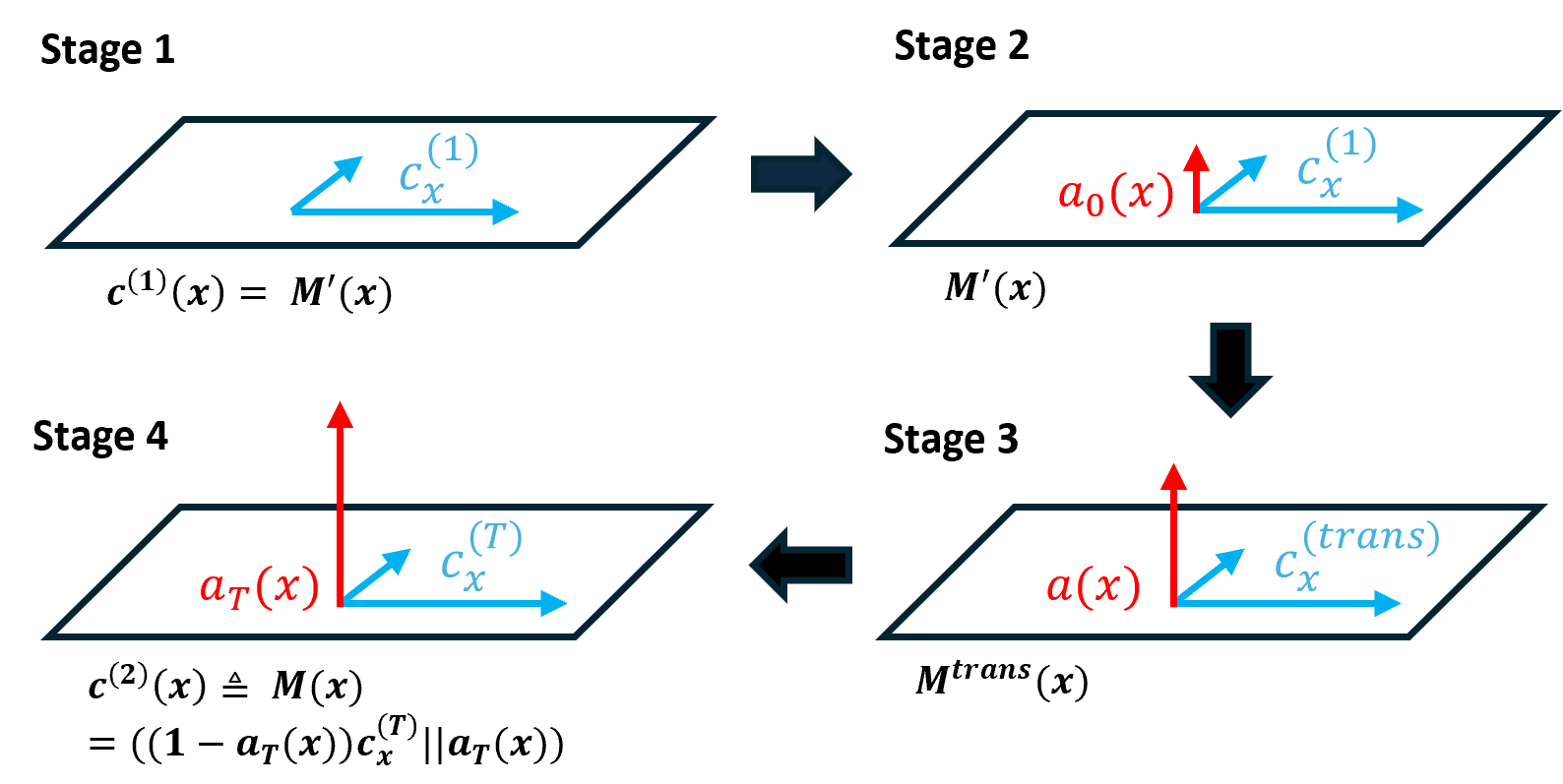}
    \caption{Inductivity in Learning}
    \label{fig:learning_process}
\end{figure}

For training a classifier using a dataset $\mathcal{D}$ with $n$ labels, given the same seed and under the convexity assumption of the objective function, learning the overall classifier with $n$ labels produces the same machine as starting from a classifier trained using $\mathcal{D}'\subseteq \mathcal{D}$ containing data points with the label being one of the first $n-1$ labels and train it further with the remaining $\mathcal{D}_n=\mathcal{D}\setminus \mathcal{D}'$ \cite{convexity_assumptions}. Here we consider the following inductive strategy. We start from $M',$ the machine that has been trained using $\mathcal{D}'.$ We can see the inductive learning from $M'$ to $M$ as embedding the original vector space of $M'(\mathcal{D}')$ to an $n$-dimensional vector space, allowing the introduction of a new basis corresponding to the additional dimension.
\begin{itemize}
    \item \textbf{Stage $1$.} For each $x\in \mathcal{X},$ denote by $\mathbf{c}^{(1)}_x= M'(x)$ is a confidence vector of length $n-1.$ 
    \item \textbf{Stage $2$.} Given $\mathbf{c}^{(1)}_x,$ we initialize a small constant $\alpha_0(x),$ and normalize $\mathbf{c}^{(1)}_x$ to have sum $1-\alpha_0(x).$ This produces a new vector $((1-\alpha_0(x))\mathbf{c}^{(1)}_x\|\alpha_0(x)) $ of length $n$ with sum of entries being equal to $1$.
    \item \textbf{Stage $3$.} Train $M$ which is a classifier with $n$ labels using $\mathcal{D}_n$ as the train data set. More specifically, $M(x)=((1-\alpha(x))\mathbf{c}^{(\mathtt{trans})}_x\|\alpha(x))$ where $\alpha(x)$ is initialized as $\alpha_0(x)$ and $\mathbf{c}^{(\mathtt{trans})}_x$ is initialized as $\mathbf{c}^{(1)}_x.$
    \item \textbf{Stage $4$.} Once the training in stage $3$ is done, we have the final values of $\alpha(x)$ and $\mathbf{c}^{(\mathtt{trans})}_x$ which we denote by $\alpha_T(x)$ and $\mathbf{c}^{(T)}_x.$ We define $\mathbf{c}^{(2)}(x)\triangleq M(x) = ((1-\alpha_T(x))\mathbf{c}^{(T)}_x\|\alpha_T(x)).$ 
\end{itemize}
An illustration of the four stages can be found in Figure \ref{fig:learning_process}.
Note that for $x\in \mathcal{D}_n,$ assuming that $M$ has a sufficiently high accuracy for input $x,$ we can see $M(x)$ as approximately orthogonal to $M'(x')$ for $x'\in \mathcal{D}'.$ Hence, in order to recover $M'(x')$ for $x'\in \mathcal{D}',$ we may consider the projection of $M(x)$ to the vector space orthogonal to $M(x)$ for $x\in \mathcal{D}_n.$

\section{Modular Projection-Redistribution Unlearning}\label{sec:proposedsol}
\begin{figure*}[t]
  \centering
  \includegraphics[width=\textwidth]{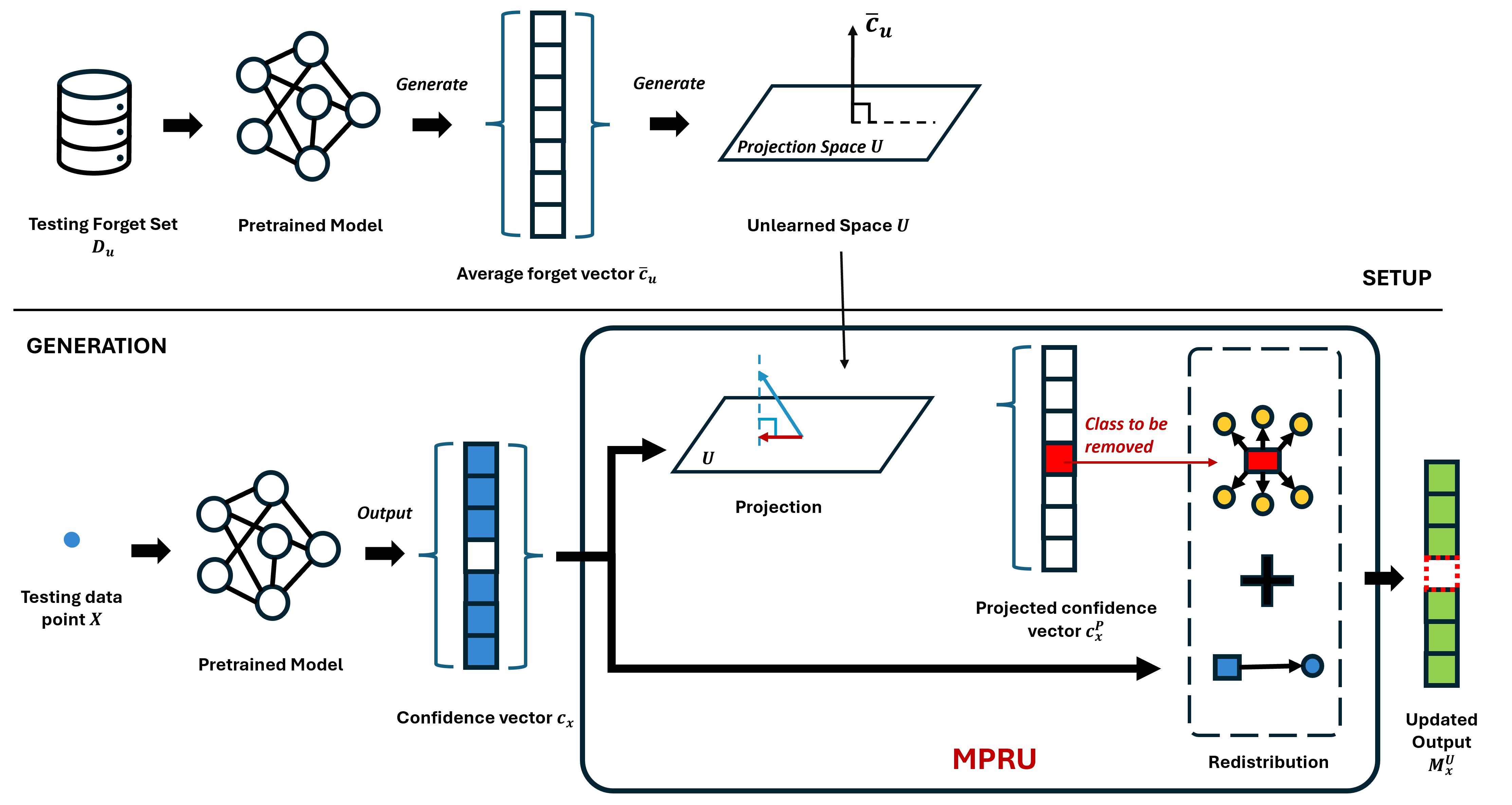}
  \caption{Overview of the proposed Modular Projection-Redistribution Unlearning (MPRU) framework}
  \label{fig:workflow}
\end{figure*}

We propose a projection and redistribution-based filter that utilizes the outputs of a well-trained model to directly obtain the prediction results under a class unlearning requirement. This filter can be implemented as a post-processing operator in model-serving systems (e.g., TensorFlow Serving, TorchServe), making it a plug-and-play data-compliance component. It operates without altering the original model, no structural modifications, retraining, or parameter tuning are involved. Conceptually, it can be viewed as the reverse of inductivity learning on single class within an established vector space, corresponding to the transition from a well pretrained $(n-1)$ classification model to an unlearned $n$ class model. Below we include Fig.\ref{fig:workflow} to show the workflow.

\subsection{Unlearning Scheme}
    

Given a pretrained model $M^P$ and label to unlearn $\mathcal{L}_u,$ Our unlearning scheme $\Pi$ defines $M^U:\mathcal{X}\rightarrow[0,1]^{n-1}$ in the following way. For $x\in \mathcal{X},$ we first define $\mathbf{c}_x=M^P(x)\in [0,1]^n.$ Here $M^U$ consists of two steps $M^U.\mathtt{Project}:[0,1]^n\rightarrow [0,1]^n$ and $M^U.\mathtt{Redistribute}:[0,1]^n\rightarrow [0,1]^{n-1}.$
\begin{itemize}
    \item $\mathbf{c}_x^P\leftarrow M^U.\mathtt{Project}(\mathbf{c}_x):$ given the original prediction $\mathbf{c}_x$ of $M^P$ from the pretrained input $x$, it projects $\mathbf{c}_x$ to the hyperplane orthogonal to $\bar{\mathbf{c}_u}\triangleq \frac{1}{N_u}\sum_{y\in \mathcal{D}_u} M^P(y)$, the average of all outputs of $M^P$ given data from the forget set. 
    \item $\mathbf{c}_x^U\leftarrow M^U.\mathtt{Redistribute}(\mathbf{c}_x^P, \mathbf{c}_x):$ given $\mathbf{c}_x^P,$ the output of $M^U.\mathtt{Project}(\mathbf{c}_x)$, it redistributes the entry corresponding to the label to be unlearned to other remaining labels, producing $\mathbf{c}_x^U,$ the output confidence vector of $M^U$ given input $x$ over label set $\mathcal{L}_r$.
\end{itemize}

The remainder of this section is divided in the following. We first discuss $M^U.\mathtt{Project}$ in Subsection \ref{sec:project}. The discussion of $M^U.\mathtt{Redistribute}$ in Section \ref{sec:Redistribute} is further separated into two sub-steps: calculate distribution ratio $\mathbf{R}_{dist}$ in Section \ref{sec:R_dist} and obtain reduction ratio $\mathbf{R}_{red}$ in Section \ref{sec:R_red}. Lastly, we put all these steps together to obtain the overall function $M^U(x)$ in Section \ref{sec:Combined}.




\subsection{Projection}\label{sec:project}
Tracing back from Step 4 to Step 1 of learning by induction, as illustrated in Section \ref{sec:inductivity}, the direction of the $n^{th}$ dimension and the original $(n-1)$-dimensional vector space are of crucial use to reverse the learning. We reckon that the introduction of a new class is equivalent to opening up a distinct subspace within the overall solution space. This subspace provides a form of isolation from others during categorization, even when overlaps exist in the feature space.


Note that from the discussion in Section \ref{sec:inductivity}, for $x\in \mathcal{D}_u$, assuming that $M$ has a sufficiently high accuracy for input $x,$ we can see $M(x)$ as approximately orthogonal to $M'(x')$ for $x'\in \mathcal{D}'.$ 

Hence, in order to recover $M'(x')$ for $x'\in \mathcal{D}',$ we may consider the projection of $M(x)$ to the vector space orthogonal to $M(x)$ for $x\in \mathcal{D}_u.$ In order to reverse the steps described in Section \ref{sec:inductivity}, first, we project $\mathbf{c}_x$ to a hyperplane that is orthogonal to the confidence vectors output by $M^P$ given data from $\mathcal{D}_u.$ Here the representation of those vectors is obtained by taking average of those vectors,
\[\bar{\mathbf{c}_u}\triangleq \frac{1}{N_u}\sum_{y\in \mathcal{D}_u} M^p(y).\]

Our aim is then to project all $\mathbf{c}_x$ for $x\in \mathcal{D}$ to the hyperplane $U$ that is orthogonal to $\bar{\mathbf{c}_u}.$ Note that in this case, by the assumption that $M^P$ has sufficiently high performance with respect to $\mathcal{D}_u,$ the variance of $M^p(y)$ of $y\in \mathcal{D}_u$ is small, which means that projecting $\mathbf{c}_y$ to $U$ will produce a very small vector.

Recall that to calculate such projection, one way is through the projection matrix $P$~\cite{Gilbert_Strang}. To achieve this, we find an orthonormal basis $\{q_1,\cdots, q_{n-1}\},$ each of length $n,$ of $U.$ Define $A,$ an $n\times n-1$ matrix with its $\ell$-th column being $q_l.$ Then projecting a vector $v$ to $U$ equals $Pv$ where $P=A\cdot A^T.$ Here $\cdot$ represents standard matrix multiplication and $T$ represents the transpose of the matrix.

Given $\mathbf{c}_x,$ further produce $\mathbf{c}_x^P = P\cdot \mathbf{c}_x.$ It is easy to see that now $\mathbf{c}_x^P\in U\subseteq \mathbb{R}^n.$ Now note that in this case, $\mathbf{c}_x^P(j),$ its $j$-th entry, corresponding to the label to be forgotten, can be seen similar to the value of $\alpha_0(x)$ described in Section \ref{sec:inductivity}.

\subsection{Redistribution}\label{sec:Redistribute}

Without loss of generality and to simplify the presentation, we assume that the label to be forgotten is the last label, which is label $n.$ Here given $\mathbf{c}_x,$ we consider the $n$-th entry, which we denote by $\mathbf{c}_{x,u}$ and the remaining vector of length $n-1,$ which we denote by $\mathbf{c}_{x,r}$. We use the same notation for any vector of length $n$ when extracting the $n$-th entry from the vector. Our strategy for building $M^U(x)$ is to distribute $\mathbf{c}_{x,u}$ to the remaining $n-1$ labels, obtaining $\mathbf{c}_{x,u}^U$ and to adjust $\mathbf{c}_{x,r},$ obtaining $\mathbf{c}_{x,r}^U,$ where we define $\mathbf{c}_x^U\triangleq  M^U(x)= \mathbf{c}_{x,u}^U+\mathbf{c}_{x,r}^U.$ First, we consider the distribution of $\mathbf{c}_{x,u}.$ 

\subsubsection{Distributing $\mathbf{c}_{x,u}$}\label{sec:R_dist}
Here recall that given $\bar{\mathbf{c}_u},$ Consider $\bar{\mathbf{c}_u}^P=P\cdot \bar{\mathbf{c}_u}$ and $\bar{\mathbf{c}_{u,r}}^P,$ its sub-vector after we remove the last entry. We can see $\bar{\mathbf{c}_{u,r}}^P$ as the effect of the forget set on the other labels. Hence, we can distribute $\mathbf{c}_{x,u}$ using the proportion defined in $\bar{\mathbf{c}_{u,r}}^P.$ Define
\[\mathbf{c}_{x,u}^U=\mathbf{c}_{x,u}\cdot\tilde{\mathbf{c}_{u,r}}^P\]
where $\tilde{\mathbf{c}_{u,r}}^P$ is the $L_1$-normalization of $\bar{\mathbf{c}_{u,r}}^P.$

\begin{figure}[h]
    \centering
    \includegraphics[width=0.48\textwidth]{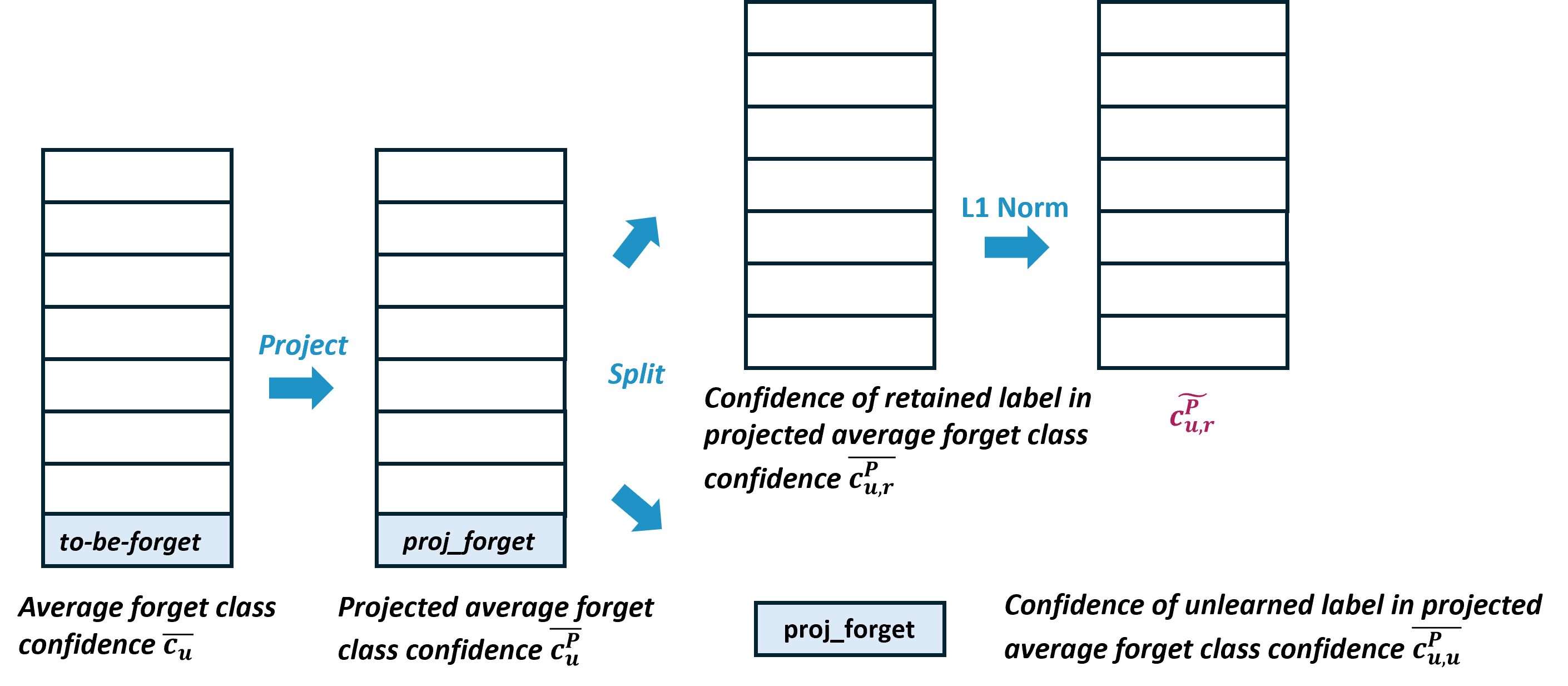}
    \caption{Distribute Factor $\tilde{\mathbf{c}_{u,r}}^P$}
    \label{fig:R_dist}
\end{figure}

\subsubsection{Adjusting $\mathbf{c}_{x,r}$}\label{sec:R_red}
Now we consider the adjustment of $\mathbf{c}_{x,r}.$ Recall that since $\mathbf{c}_x$ sums up to $1,$ we have $\mathbf{c}_{x,r}$ sums up to $1-\mathbf{c}_{x,u}.$ 

Recall that in Section \ref{sec:inductivity}, for the original classifier with $n-1$ labels, the first $n-1$ entries sum up to $(1-\alpha_0(x)).$ Here since projecting $\mathbf{c}_x$ to $U$ removes the effect of the training using data with label $n,$ we can see $\mathbf{c}_{x,u}^P$ as the value of $\alpha_0(x).$ Since we want $\mathbf{c}_{x,r}$ to sum up to $1-\alpha_0(x),$ we have

\[\mathbf{c}_{x,r}^U=\frac{1-\mathbf{c}_{x,u}^P}{1-\mathbf{c}_{x,u}}\mathbf{c}_{x,r}.\]

\subsection{Entire Unlearning Formula}\label{sec:Combined}
Combining all the discussion provided in the previous sections, for any $x\in \mathcal{X},$ noting that $\mathbf{c}_{x,u}^U+\mathbf{c}_{x,r}^U$ sums up to $\mathbf{c}_{x,u} + 1-\mathbf{c}_{x,u}^P,$ to ensure that the output sums up to $1,$ we define

\[M^U(x)=\frac{1}{\mathbf{c}_{x,u} + 1-\mathbf{c}_{x,u}^P}\left(\mathbf{c}_{x,u}\cdot \tilde{\mathbf{c}_{u,r}}^P+\frac{1-\mathbf{c}_{x,u}^P}{1-\mathbf{c}_{x,u}}\mathbf{c}_{x,r}\right).\]





\section{Asymptotic Complexity Analysis}\label{sec:complexity}
In this section, we analyse the asymptotic complexity of our proposed process. In our analysis, we assume that complexity is measured by amount of real number multiplication/ division. We further assume that real number addition has negligible complexity compared to real number multiplication. Lastly, we denote by $\mathcal{C}(M),$ the complexity of one call of the forward propagation of $M$ in terms of the number of real number multiplications. 

Based on such assumption, we note that our proposed protocol consists of the following steps:
\begin{enumerate}
    \item Calculation of $M^P(x_i)$ for $i=1,\cdots, N:$ \\
    Note that this is also required in any machine unlearning or machine learning process and it takes $N\cdot\mathcal{C}(M^P).$
    \item Calculation of $\bar{c}_u:$ \\
    This is done by averaging $N_u$ values, which takes $n$ real number division.
    \item Calculation of $\{q_1,\cdots,q_{n-1}\}$: \\
    Note that the orthonormal basis can be done via Gram-Schmidt orthogonalization that takes $O(n^3)$ multiplications
    \item Calculation of $P$: \\
    This is done by multiplying a $n\times (n-1)$ matrix by a $(n-1)\times n$ matrix. A standard matrix multiplication takes $O(n^3)$ multiplications 
    \item Calculation of $\mathbf{c}_x^P$ for $x\in \mathcal{D}:$ \\
    This takes $O(Nn^2).$
    \item For $x\in \mathcal{D},$ calculate $M^U(x)$ following the equation in Section \ref{sec:Combined}. This is simply a linear combination of vectors of length $n-1.$ Hence it takes $O(n)$ multiplications. So in total, it takes $O(Nn)$ multiplications.
\end{enumerate}

So in total, the complexity of our proposed solution is
\[\mathcal{C}=N\cdot\left(\mathcal{C}(M^P)+O(n^2)\right)+O(n^3)\] real number multiplications. In such case, since $N\gg n,$ we can simplify it to $\mathcal{C}=N\left(\mathcal{C}(M^P)+O(n^2)\right).$ Note that for any machine unlearning solution, including full retraining approaches, the computation of $M^P(x)$ for $x\in \mathcal{D}$ is always needed. Aside from such step, our complexity is $O(Nn^2).$ It is easy to see that such complexity is lower than even one iteration of step of training phase. This shows that asymptotically, our proposed solution has much smaller complexity compared to, in particular, a full retraining approach.

\section{Unlearning Experiments with CIFAR-10, CIFAR-100, and Covertype}\label{sec:experiment}

\textbf{Datasets and Models.} We evaluate MPRU on three datasets: CIFAR-10, CIFAR-100, and Covertype. 
CIFAR-10 contains 60{,}000 $32\times32$ color images in 10 classes, split into 5{,}000 training and 1{,}000 testing images per class~\cite{krizhevsky2009learning}. 
CIFAR-100 contains 60{,}000 images in 100 classes, split into 500 training and 100 testing images per class~\cite{krizhevsky2009learning}. 
Covertype is a tabular dataset with 581{,}012 instances and 54 features (10 numeric and 44 binary), split into a stratified 80/20 train/test set with seven output classes~\cite{blackard1999covertype,dua2019uci}.

We use ResNet-18 for image datasets and XGBoost for Covertype to demonstrate model agnosticism. 
Including both image and tabular data shows our method’s applicability across diverse data structures, while Covertype’s size highlights scalability. 
All models are trained with ten fixed random seeds (42, 602, 311, 637, 800, 543, 969, 122, 336, 93) for reproducibility. 
For each seed, one full model is trained on all classes, and additional models are trained with one class removed at a time; for CIFAR-100, only 40 classes are selected for removal.

Images are normalized using dataset-specific means and standard deviations. 
For CIFAR-10, mean = (0.5071, 0.4865, 0.4409) and std = (0.2673, 0.2564, 0.2762); 
for CIFAR-100, mean = (0.5071, 0.4867, 0.4408) and std = (0.2675, 0.2565, 0.2761). 
Image models are trained for 50 epochs with batch size 512 using SGD 
(momentum = 0.9, weight decay = $5\times10^{-4}$, initial learning rate = 0.1). 
Covertype models use XGBoost with 200 trees, maximum depth 6, learning rate 0.1, 
subsample ratio 0.8, column subsample ratio 0.8, and objective \texttt{multi:softprob}. 
Numeric features are standardized to zero mean and unit variance, 
and categorical features are one-hot encoded.


\textbf{Machine} All experiments are conducted in a container with the following allocated resources: an AMD EPYC 7763 CPU (2.45–3.5 GHz, 64 cores / 256 vCPUs), an NVIDIA A100 GPU with 40 GB VRAM, 64 GB DDR4 RAM, running Ubuntu 22.04 LTS (64-bit), with PyTorch 2.8 and CUDA 12.3.

\textbf{Metrics} We evaluate the performance of MPRU using four key metrics: \textit{forget/retain accuracy}, \textit{Kullback–Leibler (KL) divergence}, \textit{mean squared error (MSE)}, and \textit{average runtime}, as elaborated in Section \ref{sec:general_scheme_metrics}. 
Forget and retain accuracy are the most widely adopted indicators in machine unlearning, quantifying how effectively the model removes targeted information while preserving knowledge of non-targeted classes~\cite{rangel2024hypernetworks, bonato2024scar}. 
Accuracy alone, however, does not fully capture distributional differences in model behavior. Therefore, we complement it with KL divergence to measure the shift in output probability distributions between the retrained and redistributed models, following practices established in recent works such as~\cite{wang2025rklu, xu2025information}. 
Additionally, MSE of confidence vectors is included to provide a symmetric and scale-sensitive measure of deviation between the two models’ output probabilities, as discussed in recent surveys on unlearning techniques~\cite{mdpi2025survey}. 
Finally, average runtime is reported to evaluate computational efficiency, as highlighted by recent surveys emphasizing that practical unlearning must achieve significant time savings relative to full retraining~\cite{survey2025runtime}.

We make the implementation of our solution available at \url{https://github.com/dgunamardi/MPRU_CIFAR} and \url{https://github.com/dgunamardi/MPRU_Covertype}.

\subsection{Experimental Results and Analysis}

\begin{table*}[htbp]
\caption{Forget and Retain Accuracy for CIFAR-10 with Full Retraining and MPRU.}
\begin{center}
\begin{tabular}{|c|c|c|c|c|c|}
\hline
\textbf{unlearned}& {\textbf{Pretrained}}&{\textbf{Full Retraining}}&{\textbf{MPRU}}&\multicolumn{2}{|c|}{\textbf{Accuracy Difference}}\\
\cline{5-6} 
\textbf{class} & \textbf{\textit{Retain Acc.}} & \textbf{\textit{Retain Acc.}}& \textbf{\textit{Retain Acc.}} &\textbf{$\epsilon_p$}&\textbf{$\epsilon_r$}\\
\hline
0    & 0.9310    & 0.9224    & 0.9364   &  0.0054     &0.0140\\
1    & 0.9277    & 0.9252    & 0.9311   &  0.0034     &0.0059\\
2    & 0.9351    & 0.9330    & 0.9413   &  0.0062     &0.0083\\
3    & 0.9409    & 0.9429    & 0.9507   &  0.0098     &0.0078\\
4    & 0.9314    & 0.9283    & 0.9371   &  0.0057     &0.0088\\
5    & 0.9359    & 0.9356    & 0.9461   &  0.0102     &0.0105\\
6    & 0.9287    & 0.9250    & 0.9318   &  0.0031     &0.0068\\
7    & 0.9294    & 0.9242    & 0.9337   &  0.0043     &0.0095\\
8    & 0.9284    & 0.9189    & 0.9324   &  0.0040     &0.0135\\
9    & 0.9294    & 0.9263    & 0.9326   &  0.0032     &0.0063\\
\hline
\end{tabular}
\label{tab:ACC_CIFAR10}
\end{center}
\end{table*}

\begin{table*}[htbp]
\caption{Forget and Retain Accuracy for CIFAR-100 with full retraining and MPRU. Few unlearned classes selected for brevity. }
\begin{center}
\begin{tabular}{|c|c|c|c|c|c|}
\hline
\textbf{unlearned}& {\textbf{Pretrained}}&{\textbf{Full Retraining}}&{\textbf{MPRU}}&\multicolumn{2}{|c|}{\textbf{Accuracy Difference}}\\
\cline{5-6} 
\textbf{class} & \textbf{\textit{Retain Acc.}} & \textbf{\textit{Retain Acc.}}& \textbf{\textit{Retain Acc.}} &\textbf{$\epsilon_p$}&\textbf{$\epsilon_r$}\\
\hline
11   &  0.7384 & 0.7388     & 0.7409    & 0.0025   &0.0021    \\
23   &  0.7353 & 0.7361     & 0.7362    & 0.0009   &0.0001    \\
35   &  0.7383 & 0.7432     & 0.7417    & 0.0034   &0.0015    \\
49   &  0.7347 & 0.7325     & 0.7358    & 0.0010   &0.0033     \\
53   &  0.7345 & 0.7362     & 0.7353    & 0.0007   &0.0009    \\
61   &  0.7362 & 0.7342     & 0.7374    & 0.0012   &0.0032    \\
68   &  0.7342 & 0.7294     & 0.7347    & 0.0005   &0.0053    \\
72   &  0.7389 & 0.7407     & 0.7411    & 0.0022   &0.0004    \\
88   &  0.7354 & 0.7315     & 0.7364    & 0.0010   &0.0049     \\
97   &  0.7360 & 0.7322     & 0.7370    & 0.0010   &0.0048     \\
\hline
\end{tabular}
\label{tab:ACC_CIFAR100}
\end{center}
\end{table*}

\begin{table*}[htbp]
\caption{Forget and Retain Accuracy for Covertype with Full Retraining and MPRU.}
\begin{center}
\begin{tabular}{|c|c|c|c|c|c|}
\hline
\textbf{unlearned}& {\textbf{Pretrained}}&{\textbf{Full Retraining}}&{\textbf{MPRU}}&\multicolumn{2}{|c|}{\textbf{Accuracy Difference}}\\
\cline{5-6} 
\textbf{class} & \textbf{\textit{Retain Acc.}} & \textbf{\textit{Retain Acc.}}& \textbf{\textit{Retain Acc.}} &\textbf{$\epsilon_p$}&\textbf{$\epsilon_r$}\\
\hline

0   & 0.8638    & 0.9598    & 0.9471    & 0.0834    & 0.0126 \\
1   & 0.8121    & 0.9677    & 0.9426    & 0.1306    & 0.0250 \\
2   & 0.8418    & 0.8560    & 0.8492    & 0.0074    & 0.0068 \\
3   & 0.8450    & 0.8468    & 0.8453    & 0.0003    & 0.0015 \\
4   & 0.8508    & 0.8543    & 0.8513    & 0.0005    & 0.0029 \\
5   & 0.8486    & 0.8559    & 0.8526    & 0.0040    & 0.0034 \\
6   & 0.8440    & 0.8475    & 0.8461    & 0.0021    & 0.0014 \\
\hline
\end{tabular}
\label{tab:ACC_Covertype}
\end{center}
\end{table*}

\begin{table*}[htbp]
\caption{Three way KL Divergence for CIFAR-10 with Pretrained, Full Retraining, and MPRU.}
\centering
\label{tab:KL_CIFAR10}
\begin{tabular}{|c|c|c|c|c|c|c|}
\hline
\textbf{unlearned}&\multicolumn{2}{|c|}{\textbf{Pretrained - MPRU}}&\multicolumn{2}{|c|}{\textbf{Pretrained - Retrain}}&\multicolumn{2}{|c|}{\textbf{Retrain - MPRU}} \\
\cline{2-7} 
\textbf{class} & \textbf{\textit{Retain KL}}& \textbf{\textit{Forget KL}}& \textbf{\textit{Retain KL}}& \textbf{\textit{Forget KL}}& \textbf{\textit{Retain KL}}& \textbf{\textit{Forget KL}} \\
\hline
0 & 0.2527 & 25.7242 & 0.2848 & 10.2925 & 0.2051 & 0.7853 \\
1 & 0.1143 & 26.5457 & 0.2576 & 14.3729 & 0.2238 & 0.9495 \\
2 & 0.2458 & 24.4400 & 0.2676 & 10.4631 & 0.1530 & 0.8139 \\
3 & 0.4680 & 22.9906 & 0.3396 & 10.8121 & 0.1260 & 0.7170 \\
4 & 0.2672 & 25.6531 & 0.2878 & 10.5195 & 0.1758 & 0.9158 \\
5 & 0.3618 & 24.3058 & 0.3440 & 14.3450 & 0.1394 & 0.7958 \\
6 & 0.1720 & 26.0013 & 0.2616 & 12.6089 & 0.1761 & 0.9037 \\
7 & 0.1302 & 26.0764 & 0.2423 & 11.6620 & 0.1994 & 0.8600 \\
8 & 0.1300 & 26.1147 & 0.2347 & 11.6371 & 0.1804 & 0.8558 \\
9 & 0.1635 & 26.2583 & 0.2573 & 12.0922 & 0.1826 & 0.6744 \\
\hline
\end{tabular}
\centering
\end{table*}

\begin{table*}[htbp]
\caption{Three way KL Divergence for CIFAR-100 with Pretrained, Full Retraining, and MPRU. Few unlearned classes selected for brevity.}
\centering
\label{tab:KL_CIFAR100}
\begin{tabular}{|c|c|c|c|c|c|c|}
\hline
\textbf{unlearned}&\multicolumn{2}{|c|}{\textbf{Pretrained - MPRU}}&\multicolumn{2}{|c|}{\textbf{Pretrained - Retrain}}&\multicolumn{2}{|c|}{\textbf{Retrain - MPRU}} \\
\cline{2-7} 
\textbf{class} & \textbf{\textit{Retain KL}}& \textbf{\textit{Forget KL}}& \textbf{\textit{Retain KL}}& \textbf{\textit{Forget KL}}& \textbf{\textit{Retain KL}}& \textbf{\textit{Forget KL}} \\
\hline
11 & 0.1322 & 12.0151 & 0.6095 & 6.7543  & 0.5412 & 0.8350 \\
24 & 0.0418 & 21.8199 & 0.5794 & 11.6765 & 0.5516 & 0.7668 \\
23 & 0.0510 & 21.4887 & 0.5824 & 9.3599  & 0.5646 & 1.1756 \\
30 & 0.0815 & 17.2617 & 0.5826 & 9.3178  & 0.5394 & 0.8679 \\
35 & 0.1166 & 12.5886 & 0.6064 & 6.8333  & 0.5448 & 0.7015 \\
48 & 0.0455 & 24.7506 & 0.5875 & 11.7298 & 0.5559 & 0.8908 \\
59 & 0.0967 & 18.2538 & 0.5958 & 8.1140  & 0.5451 & 0.7609 \\
61 & 0.0745 & 18.0739 & 0.6007 & 8.8842  & 0.5542 & 0.8477 \\
68 & 0.0336 & 25.0891 & 0.5755 & 11.3400 & 0.5586 & 0.6965 \\
71 & 0.0631 & 22.1916 & 0.6102 & 10.5669 & 0.5750 & 0.7891 \\
92 & 0.0761 & 17.4311 & 0.6052 & 8.2614  & 0.5608 & 0.8820 \\
\hline
\end{tabular}
\centering
\end{table*}

\begin{table*}[htbp]
\caption{Three way KL Divergence for Covertype with Pretrained, Full Retraining, and MPRU.}
\centering
\label{tab:KL_Covertype}
\begin{tabular}{|c|c|c|c|c|c|c|}
\hline
\textbf{unlearned}&\multicolumn{2}{|c|}{\textbf{Pretrained - MPRU}}&\multicolumn{2}{|c|}{\textbf{Pretrained - Retrain}}&\multicolumn{2}{|c|}{\textbf{Retrain - MPRU}} \\
\cline{2-7} 
\textbf{class} & \textbf{\textit{Retain KL}}& \textbf{\textit{Forget KL}}& \textbf{\textit{Retain KL}}& \textbf{\textit{Forget KL}}& \textbf{\textit{Retain KL}}& \textbf{\textit{Forget KL}} \\
\hline
0 & 4.2294 & 19.2972 & 4.2238 & 19.2794 & 0.0279 & 0.2776 \\
1 & 5.7677 & 20.8186 & 5.7452 & 20.7871 & 0.0659 & 0.3176 \\
2 & 0.3899 & 21.0516 & 0.3931 & 21.0230 & 0.0105 & 0.3820 \\
3 & 0.0239 & 21.2772 & 0.0285 & 21.2693 & 0.0047 & 0.1448 \\
4 & 0.2132 & 12.0318 & 0.2177 & 12.0226 & 0.0053 & 0.0513 \\
5 & 0.3273 & 15.4836 & 0.3304 & 15.4394 & 0.0079 & 0.2179 \\
6 & 0.1904 & 21.3989 & 0.1958 & 21.3944 & 0.0056 & 0.0738 \\
\hline
\end{tabular}
\centering
\end{table*}


\begin{table*}[htbp]
\caption{MSE for CIFAR-10 between full retraining and MPRU.}
\begin{center}
\begin{tabular}{|c|c|c|c|c|}
\hline
\textbf{unlearned}&\multicolumn{4}{|c|}{\textbf{MSE}} \\
\cline{2-5} 
\textbf{class} & \textbf{\textit{Mean}}&  \textbf{\textit{Std}}& \textbf{\textit{Max}}& \textbf{\textit{$\%$ of data $<$ mean}}\\
\hline
0 & 0.011053380  & 0.031603076   & 0.19923480 & 84.76\\
1 & 0.012479043  & 0.032833830   & 0.19970553 & 83.57\\
2 & 0.009676244  & 0.029312728   & 0.19834545 & 85.27\\
3 & 0.009253293  & 0.029170103   & 0.19815752 & 86.62\\
4 & 0.011066727  & 0.032002673   & 0.19950497 & 84.96\\
5 & 0.009979398  & 0.030091710   & 0.19785269 & 86.22\\
6 & 0.011209808  & 0.031188045   & 0.19914520 & 83.66\\
7 & 0.010483603  & 0.030038733   & 0.19702692 & 84.71\\
8 & 0.011447178  & 0.031061988   & 0.19823640 & 83.44\\
9 & 0.011096914  & 0.030438000   & 0.19871760 & 83.87\\
\hline
\end{tabular}
\label{tab:MSE_CIFAR10}
\end{center}
\end{table*}

\begin{table*}[htbp]
\caption{MSE for CIFAR-100 between full retraining and MPRU.}
\begin{center}
\begin{tabular}{|c|c|c|c|c|}
\hline
\textbf{unlearned}&\multicolumn{4}{|c|}{\textbf{MSE}} \\
\cline{2-5} 
\textbf{class} & \textbf{\textit{Mean}}& \textbf{\textit{Std}}& \textbf{\textit{Max}}& \textbf{\textit{$\%$ of data $<$ mean}}\\
\hline
14 & 0.0016582306 & 0.0030411375 & 0.019123511 & 72.86\\
23 & 0.0016798169 & 0.0030739666 & 0.019517267 & 73.19\\
35 & 0.0015975103 & 0.0029158657 & 0.019220762 & 72.80\\
49 & 0.0017007544 & 0.0030837378 & 0.019835899 & 72.73\\
53 & 0.0016764778 & 0.0030733860 & 0.019493477 & 73.07\\
61 & 0.0016641258 & 0.0030513064 & 0.019663265 & 72.72\\
68 & 0.0016342413 & 0.0030319234 & 0.019161247 & 73.19\\
72 & 0.0016502724 & 0.0030256314 & 0.019197830 & 73.21\\
88 & 0.0016507783 & 0.0030212298 & 0.019586952 & 72.99\\
97 & 0.0016627447 & 0.0030373225 & 0.019559670 & 72.92\\
\hline
\end{tabular}
\label{tab:MSE_CIFAR100}
\end{center}
\end{table*}

\begin{table*}[htbp]
\caption{MSE for Covertype between full retraining and MPRU.}
\begin{center}
\begin{tabular}{|c|c|c|c|c|}
\hline
\textbf{unlearned}&\multicolumn{4}{|c|}{\textbf{MSE}} \\
\cline{2-5} 
\textbf{class} & \textbf{\textit{Mean}}& \textbf{\textit{Std}}& \textbf{\textit{Max}}& \textbf{\textit{$\%$ of data $<$ mean}}\\
\hline
0 & 0.0097757099650 & 0.037428907150 & 0.24190562770 &91.69 \\
1 & 0.0126125273700 & 0.034704564670 & 0.22815240870 &86.74 \\
2 & 0.0025384519170 & 0.010328770690 & 0.18842086520 &88.21 \\
3 & 0.0004580792273 & 0.001489933060 & 0.06272989792 &79.55 \\
4 & 0.0004505896739 & 0.001494483561 & 0.12059238390 &78.35 \\
5 & 0.0010432015200 & 0.004543604681 & 0.11797128650 &85.62 \\
6 & 0.0006251359668 & 0.004649809219 & 0.22232149970 &80.41 \\
\hline
\end{tabular}
\label{tab:MSE_Covertype}
\end{center}
\end{table*}


\begin{table*}[htbp]
\caption{Runtime comparison for CIFAR-10, CIFAR-100, and Covertype with full retraining and MPRU.}
\begin{center}
\begin{tabular}{|c|c|c|}
\hline
\textbf{Dataset}&\textbf{Retraining Avg} &\textbf{MPRU Avg} \\
\textbf{} & \textbf{\textit{Runtime (s)}}& \textbf{\textit{Runtime (s)}} \\
\hline
CIFAR-10   & 200.66    & 0.00882\\
CIFAR-100  & 442.70     & 0.08412\\
Covertype  & 12.84     & 0.01698\\
\hline
\end{tabular}
\label{tab:speed}
\end{center}
\end{table*}

\begin{figure}[htbp]
    \centering
    \includegraphics[width=1\linewidth]{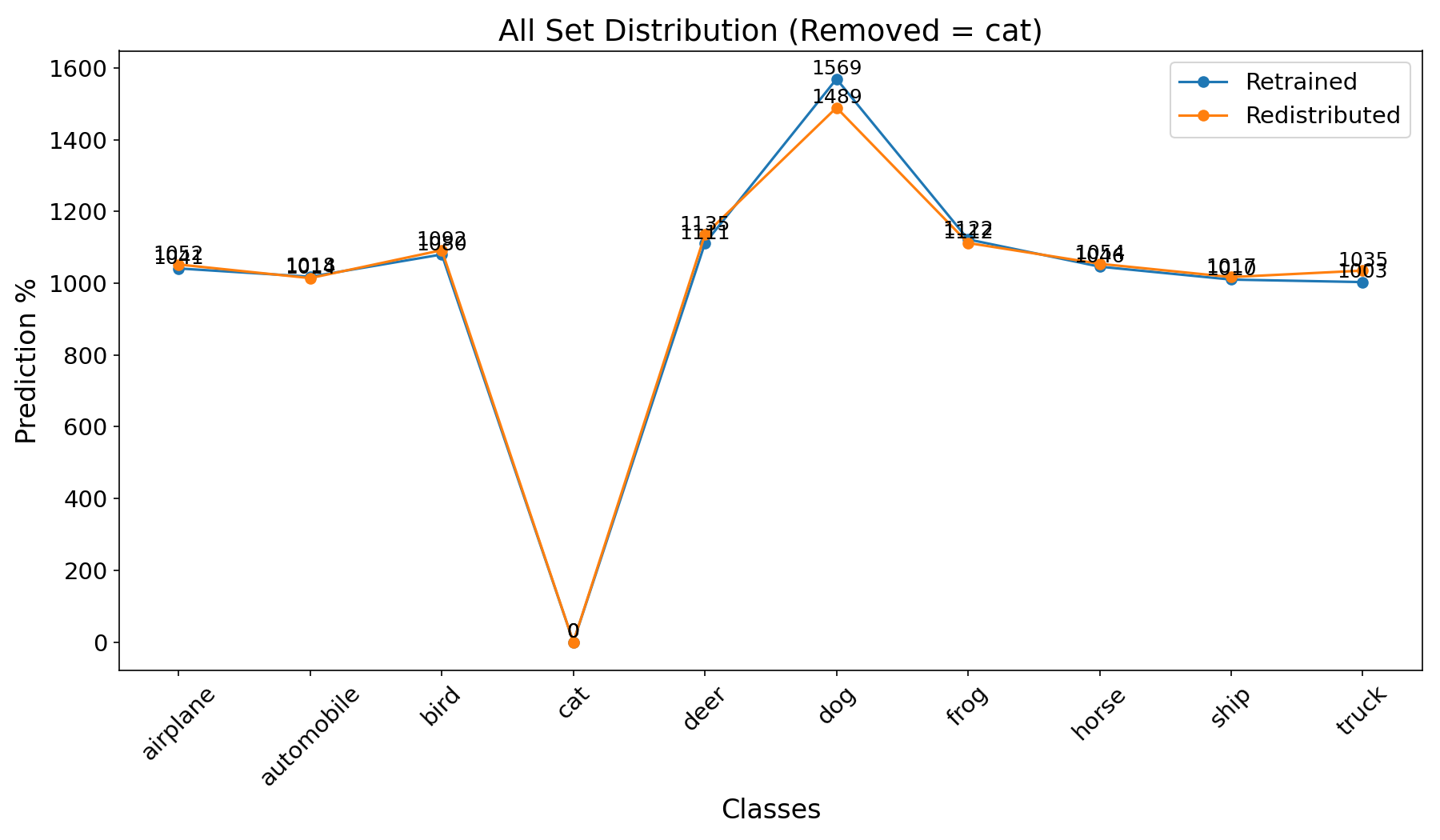}
    \caption{Prediction Comparison for CIFAR-10 between Full Retraining and MPRU with removed class 3 "cat".}
    \label{fig:cifar10_pred_cat}
\end{figure}



\begin{figure}[htbp]
    \centering
    \includegraphics[width=1\linewidth]{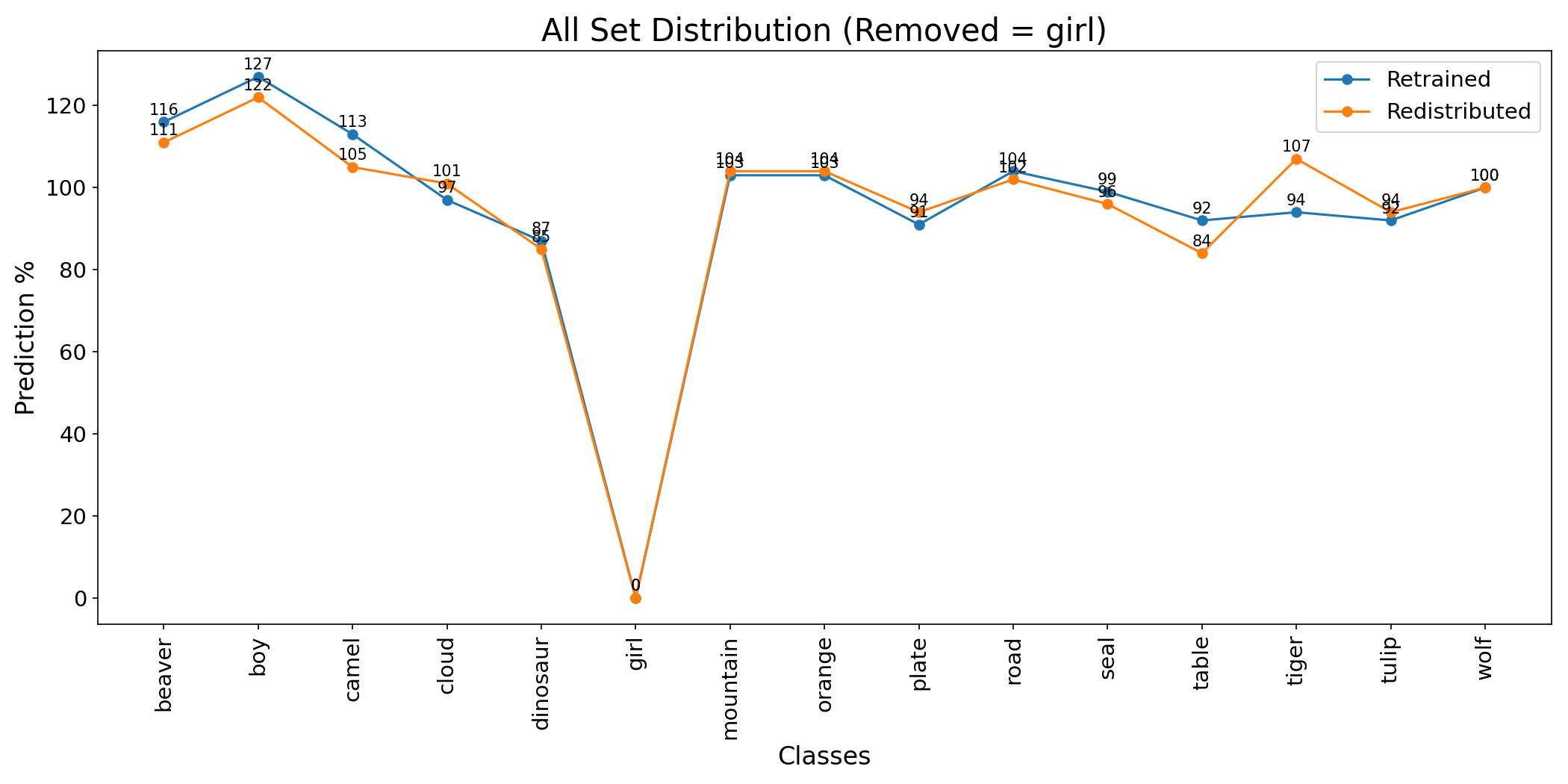}
    \caption{Prediction Comparison for CIFAR-100 between Full Retraining and MPRU with removed class 35 "girl". Select few classes are shown for brevity.}
    \label{fig:cifar100_pred_girl}
\end{figure}


\begin{figure}[htbp]
    \centering
    \includegraphics[width=1\linewidth]{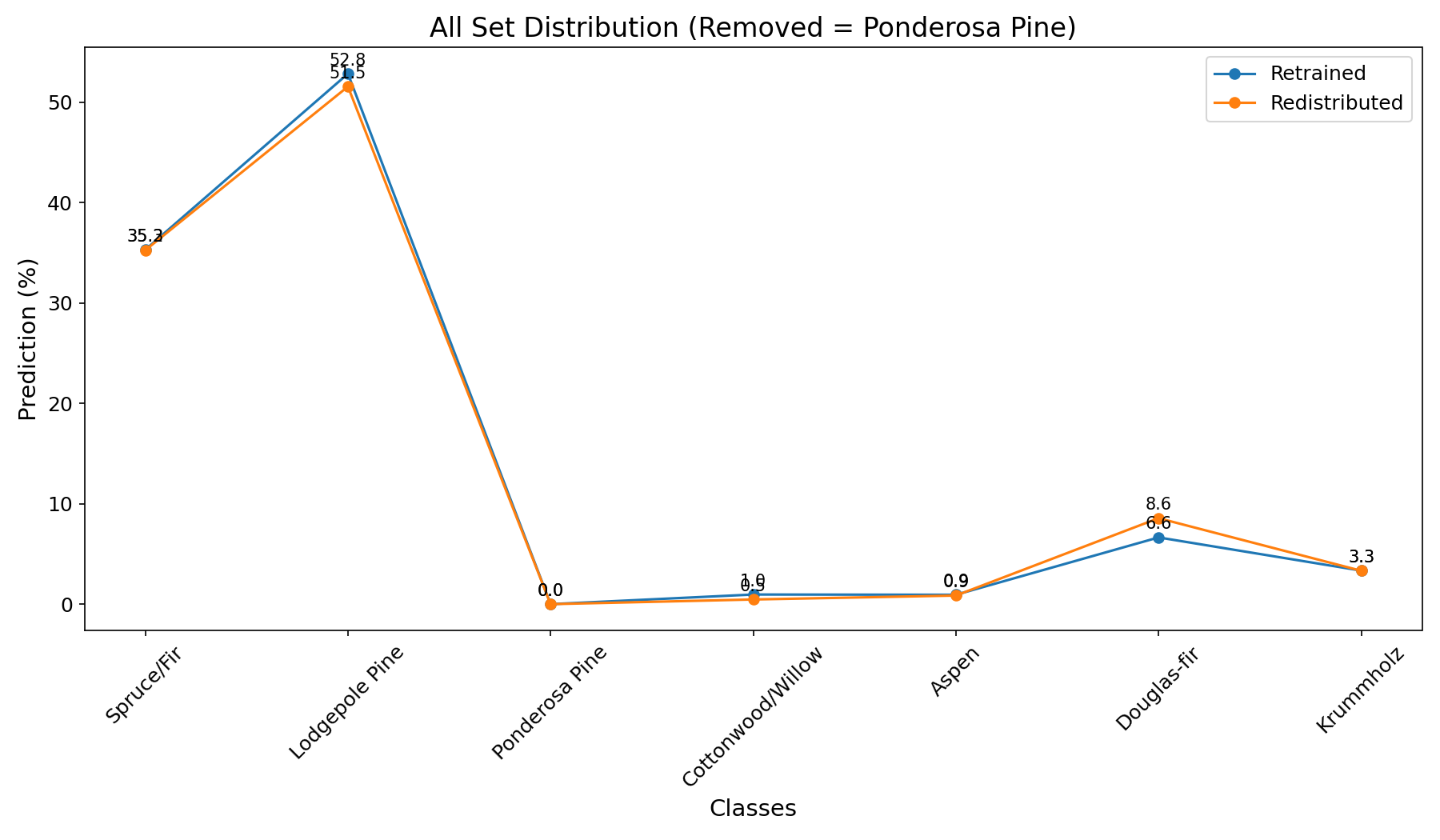}
    \caption{Prediction Comparison for Covertype between Full Retraining and MPRU with removed class 2 "Ponderosa Pine".}
    \label{fig:covertype_pred_ponderosa_pine}
\end{figure}


\input{Addition/Experiment}

\section{Conclusion}
We considered an inductive approach as a different direction to machine unlearning. This reformulates classification models into systems that are reversible and unlearning is done using our projection-redistribution framework we call MPRU. Requiring only model outputs, it is deployable as a lightweight, modular, and model-agnostic filter that produces similar outputs to retraining approaches with less complexity. Its applicability and scalability are also demonstrated by our experiments using 
image and large tabular datasets on CNN-based and tree-based models. Further, results across multiple random fixed seeds show the ability to consistently achieve the result from the gold standard of retraining.

\section*{AI-Generated Content Acknowledgment}

We acknowledge the use of ChatGPT (GPT-5, OpenAI, 2025) to assist in enhancing the clarity and readability of the manuscript. The tool was also employed to support code generation and documentation refinement for our experiments, with all outputs subsequently validated and verified by the authors. All substantive ideas, results, and conclusions presented in this work are entirely the authors’ own.

\bibliographystyle{IEEEtran}
\bibliography{ref}

\end{document}

%% file: Addition/Experiment.tex
In this section, we present our experimental result on the metrics described previously.
\subsubsection{Forget/Retain Accuracy}
First, we consider the most widely adopted indicators in machine unlearning. Tables \ref{tab:ACC_CIFAR10},  \ref{tab:ACC_CIFAR100}, and \ref{tab:ACC_Covertype} show the accuracy comparison between the predictions of the pretrained model, retrained model, and the prediction of MPRU with respect to both the retain set and forget set for the three datasets respectively. Here we consider various cases depending on the class to be unlearned. We note that since in both retraining and MPRU, it is impossible for it to output any probability for the forgotten label, clearly, the accuracy for the forgotten label is $0$ for all cases. So we are more interested in the accuracy change for the retained labels.

We observe that the accuracy of MPRU on the retain set is very similar to both the accuracy level of the pretrained model and the fully retrained model in any scenario. The evaluation of the forget set is further analyzed by examining the distribution of forget samples predicted as retained classes by the unlearned model. This can be observed from illustrations of prediction distribution graph, which can be found in Figure \ref{fig:cifar10_pred_cat} for CIFAR-10, Figure~\ref{fig:cifar100_pred_girl} for CIFAR-100, and Figure~\ref{fig:covertype_pred_ponderosa_pine} for Covertype.


This shows that MPRU mimics the fully retrained model while maintaining the performance of the pretrained model in terms of its accuracy. We further note that in some cases, the retain accuracy of MPRU is in fact higher than the fully retrained model. This may be seen from the fact that in fully retrained model, all training data points from the forget set are discarded. While in MPRU, information that the model has learned from such forget set is used to further refine the prediction of other classes. However, in some cases, this refinement actually causes an adverse effect on the accuracy of the retained set (see, for example, the cases when the unlearned class is $35$ or $53$ for CIFAR-100 or the cases for Covertype). In such cases, this may be caused by the fact that by redistributing such information, it blurs the separation between other classes, causing the accuracy to drop. 


Following our definition of accuracy differences from $M^U$ to $M^R$ and $M^P,$ we observe the following thresholds: $\epsilon_r\sim0.0364$ and $\epsilon_p\sim0.0136$ for CIFAR-10, implying $M^U$ achieves $0.0364$-accuracy difference to $M^R$ and $0.0136$-accuracy difference to $M^P$. $\epsilon_r\sim0.0127$ and $\epsilon_p\sim0.0042$ for CIFAR-100, implying $M^U$ achieves $0.0127$-accuracy difference to $M^R$ and $0.0042$-accuracy difference to $M^P$. $\epsilon_r\sim0.0272$ and $\epsilon_p\sim0.1322$ for Covertype, implying  $M^U$ achieves $0.0272$-accuracy difference to $M^R$ and $0.1322$-accuracy difference to $M^P$.


\subsubsection{Similarity of Confidence Vector}
We next consider a different similarity requirement that considers the more general case of the confidence vector output by the models instead of just the label with the highest confidence as what is done in the accuracy measurement. This provides a more thorough verification of the similarity of the behaviors between $M^R$ and $M^U$ in all available data points. Tables \ref{tab:KL_CIFAR10}, \ref{tab:KL_CIFAR100}, and \ref{tab:KL_Covertype} show the KL divergence average between any pair of the outputs of the pretrained model, retrained model, and the output of MPRU, with respect to both the retain set and forget set for the three datasets respectively. Here we consider various cases depending on the class to be unlearned. It is easy to see that in all scenarios, both the average KL divergence for retain set and forget set between the output of fully retrained model and MPRU are very small. More specifically, we are statistically within $(0.3595,1.2823), (0.5968,1.9202),$ and $(0.0659,0.3867)$ to $M^R$ for CIFAR-10, CIFAR-100, and Covertype respectively. The small maximum KL divergence throughout the classes for the three datasets in both retain and forget sets suggests that the $M^U$ indeed mimics the behaviour of $M^R$ in all the data points, including those data points with the forgotten label.

Next, we consider the MSE-based distance between the confidence vector output by $M^R$ and $M^U.$ This measures the similarity between the behaviors of $M^R$ and $M^U$. Even it has been done using KL divergence as described above, we believe that the MSE measurement may be interpreted more easily, providing more insights on the similarity between $M^R$ and $M^U.$ As observed in Tables \ref{tab:MSE_CIFAR10}, \ref{tab:MSE_CIFAR100}, and \ref{tab:MSE_Covertype}, for the three datasets, although $\epsilon_u$ is approximately 0.2 for CIFAR-10, 0.02 for CIFAR-100 and 0.244 for Covertype, we can see that the average MSE is much smaller, showing that in majority of the case, the MSE between the confidence vectors of $M^R$ and $M^U$ is much smaller than its upper bound. We further notice that the average of the MSE for the three datasets are much smaller with similarly small standard deviation. It can be confirmed on the last column that at least $72\%$ of the data points have MSE of at most $0.0145, 0.0017, 0.013$(mean value) respectively for the three datasets, showing that our proposed solution behaves in a much more similar manner to the baseline $M^R$ in at least $72\%$ of the time.


\subsubsection{Average Runtime}
Lastly, we consider the comparison between the runtime required for full retraining against MPRU. Clearly, due to the lack of model training step, MPRU shows a much smaller average runtime, confirming the result in complexity analysis in Section \ref{sec:complexity}.